\documentclass[10pt,twocolumn,letterpaper]{article}
\PassOptionsToPackage{numbers,sort,compress}{natbib}
\usepackage[pagenumbers]{cvpr} 

\usepackage[numbers]{natbib}
\usepackage{multicol}
\usepackage{bm}

\usepackage{twemojis}
\usepackage{utfsym}

\usepackage{times}

\usepackage{graphicx}
\usepackage{amsmath}
\usepackage{amssymb}
\usepackage{booktabs}
\usepackage{cuted}
\usepackage{duckuments}
\usepackage{algorithm}
\usepackage{algcompatible}
\usepackage{amsmath}

\definecolor{cvprblue}{rgb}{0.21,0.49,0.74}
\usepackage[pagebackref,breaklinks,colorlinks,citecolor=cvprblue]{hyperref}

\usepackage[capitalize]{cleveref}
\Crefname{section}{Sec.}{Secs.}
\crefname{section}{Sec.}{Secs.}
\Crefname{table}{Tab.}{Tabs.}
\crefname{table}{Tab.}{Tabs.}
\Crefname{figure}{Fig.}{Figs.}
\crefname{figure}{Fig.}{Figs.}
\Crefname{appendix}{App.}{Apps.}
\crefname{appendix}{App.}{Apps.}

\usepackage[utf8]{inputenc} 
\usepackage[T1]{fontenc}    
\usepackage{url}            
\usepackage{amsfonts}       
\usepackage{nicefrac}       
\usepackage[nopatch=eqnum]{microtype}      

\usepackage{epsfig}
\usepackage{tabularx}
\usepackage{bbm}
\usepackage{color}
\usepackage{wrapfig}
\usepackage{subcaption}
\usepackage{float}
\usepackage{makecell}
\usepackage[percent]{overpic}
\usepackage{adjustbox}

\usepackage{multirow}
\usepackage{inconsolata}
\usepackage{comment}
\usepackage{xspace}

\usepackage{listings}

\newcommand{\stdhide}[1]{}

\newcolumntype{Y}{>{\centering\arraybackslash}X}

\def\paperID{7381} 
\def\confName{CVPR}
\def\confYear{2025}

\title{PhysGen3D: Crafting a Miniature Interactive World from a Single Image}

\author{
    Boyuan Chen$^{1}$, Hanxiao Jiang$^{2,3}$, Shaowei Liu$^{2}$,\\
    Saurabh Gupta$^{2}$, Yunzhu Li$^{3}$, Hao Zhao$^{1}$, Shenlong Wang$^{2}$\\
    \small{$^{1}$Tsinghua University, $^{2}$University of Illinois Urbana-Champaign, $^{3}$Columbia University}
}

\begin{document}
\maketitle
\begin{strip}
    \centering
    \vspace{-30pt}
    \includegraphics[width=\linewidth]{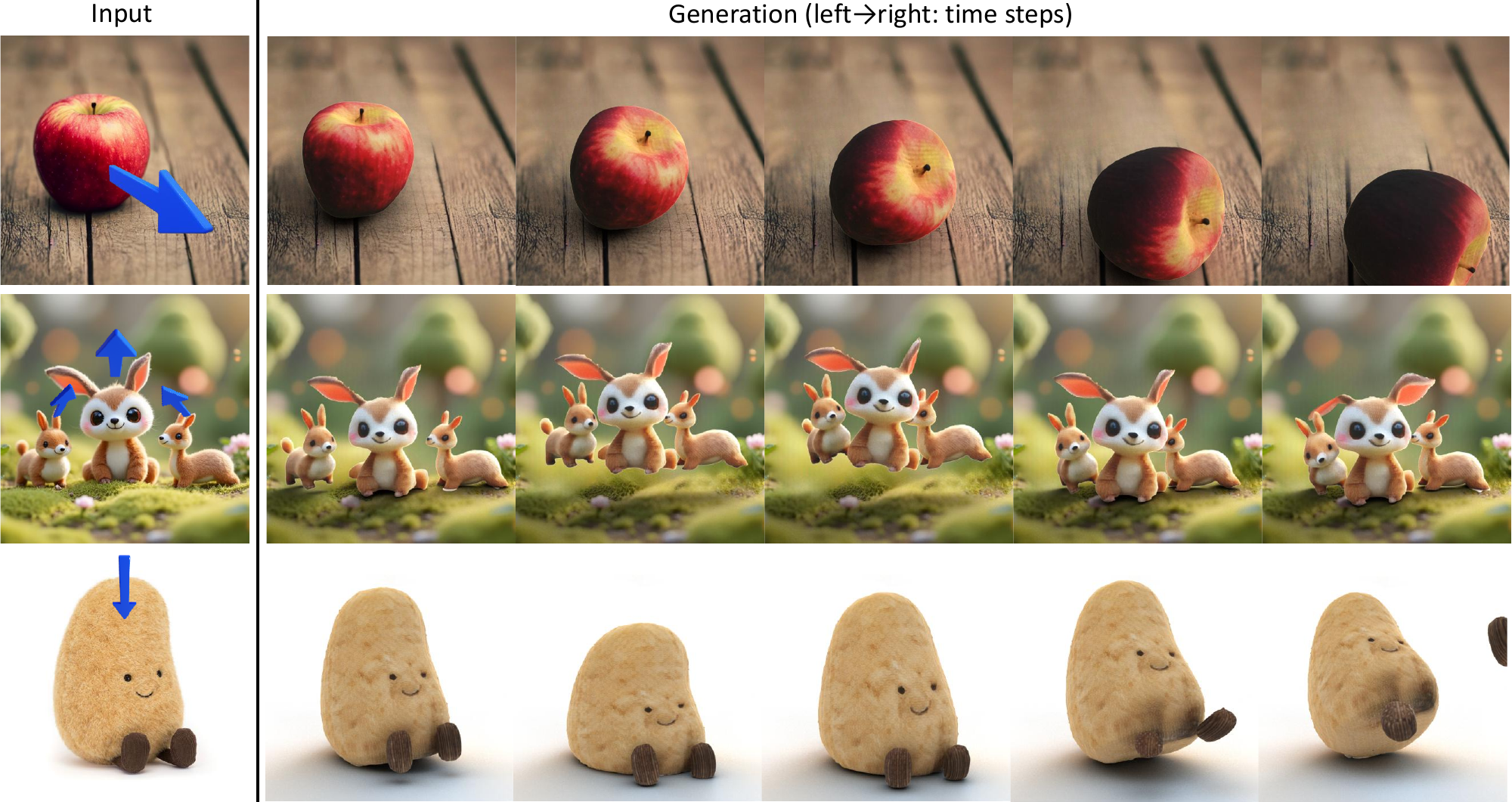}
    \captionof{figure}{
    PhysGen3D generates realistic, physically plausible motion from a single image and a text prompt by reasoning about geometry, semantics, and material properties. (a) An apple rolls under the influence of its initial velocity, friction, and shape, producing a natural progression over time. (b) Three animal figures interact dynamically, colliding after being propelled upwards and forwards. (c) A toy potato bounces back with soft-body dynamics in response to an initial downward force, capturing material-specific behaviors. PhysGen3D lets users quickly explore physics‑driven object interactions and behaviors in a compact virtual scene generated from a single input image.
    }
    \vspace{-3pt}
    \label{fig:teaser}
\end{strip}

\begin{abstract}
\vspace{-3pt}
Envisioning physically plausible outcomes from a single image requires a deep understanding of the world's dynamics. To address this, we introduce PhysGen3D, a novel framework that transforms a single image into an amodal, camera-centric, interactive 3D scene. By combining advanced image-based geometric and semantic understanding with physics-based simulation, PhysGen3D creates an interactive 3D world from a static image, enabling us to "imagine" and simulate future scenarios based on user input. At its core, PhysGen3D estimates 3D shapes, poses, physical and lighting properties of objects, thereby capturing essential physical attributes that drive realistic object interactions. This framework allows users to specify precise initial conditions, such as object speed or material properties, for enhanced control over generated video outcomes. We evaluate PhysGen3D's performance against closed-source state-of-the-art (SOTA) image-to-video models, including Pika, Kling, and Gen-3, showing PhysGen3D's capacity to generate videos with realistic physics while offering greater flexibility and fine-grained control. 
Our results show that PhysGen3D achieves a unique balance of photorealism, physical plausibility, and user-driven interactivity, opening new possibilities for generating dynamic, physics-grounded video from an image. Project page: \url{https://by-luckk.github.io/PhysGen3D}.



\end{abstract}    
\vspace{-20pt}
\section{Introduction}
\label{sec:intro}

Photographs capture snapshots of our physical world, preserving specific moments in time but leaving out the alternative outcomes that could have unfolded. 
For instance, looking at a photo, we might wonder, “What if I poke the apples to make them roll across the ground?” or “What if I squeeze the three stuffed animals closer together?” or “What if I drop my cute potato toys onto the floor?”
Humans intuitively understand how these scenarios would play out because we have an innate sense of the physical world beyond what we see in a single image. We develop a computational model that can answer such "what-if" questions by generating video outcomes from a single static image.

A promising approach toward this goal is data-driven image-to-video (I2V) generation~\cite{chuang2005animating, davis2015image, xue2018visual, holynski2021animating, schodl2000video, Szummer96a, wei2000fast}. I2V leverages diffusion-based generative models trained on vast datasets of internet images and videos, enabling the production of photorealistic videos with remarkable detail. However, I2V still has limitations in precise control and lacks 
physical grounding. 
As a result, users cannot interact freely and accurately 
to achieve specific physical effects, nor can I2V guarantee physical realism. 

On the other hand, recent research has focused on modeling the physical world from visual inputs to create digital twins, allowing for precise interactions \cite{xie2024physgaussian, zhang2025physdreamer, li2023pac, zhong20233d}. These approaches can generate virtual scenes with convincing physical interactions, but they typically require complete 3D scans from multi-view images or depth sensors, making them data-intensive. 
While some methods \cite{liu2025physgen, li2023generative, sugimoto2022water} enable interaction with a single image, they are often limited by physical constraints (e.g., rigid bodies or springs), specific object types (e.g., waterfalls), or a 2D scope.
This gap highlights the need for a generic, controllable, physically grounded, and photorealistic approach to generate video from a single image while maintaining physical realism.

In this work, we introduce PhysGen3D, a novel framework that transforms a single image into an amodal, camera-centric, interactive 3D scene, enabling realistic simulation and rendering.
Our approach combines the strengths of image-based geometric and semantic understanding~\cite{xu2024instantmesh,wang2024dust3r,ren2024grounded,suvorov2021resolution,kirillov2023segany} with physics-based simulation \cite{hu2019taichi,hu2019difftaichi,hu2021quantaichi}. 
At its core, PhysGen3D is a digital twinning method that estimates an object’s 3D shape, pose, physical and lighting properties, infers background geometry and appearance, deduces physical characteristics, and performs dimensional analysis—all from a single input image. This task is conventionally challenging due to its inherently ill-posed nature. To tackle this, we leverage various pretrained vision models, integrating their outputs to create an image-centric digital twin.

For physical simulation, we employ material point methods \cite{hu2018moving,jiang2016material}, a robust point-voxel-based framework that models counterfactual physical behaviors of objects in the image. 
Through precise inference of physical properties, simulations in the PhysGen3D environment achieve a high degree of realism and stability. We further enhance realism by applying physics-based rendering, seamlessly integrating dynamic effects back into the original image. PhysGen3D produces results that are not only visually realistic in terms of dynamics and lighting but also highly controllable, allowing users to specify initial conditions like speed and material properties. Due to the use of large pretrained models, our pipeline operates effectively without task-specific training.

Our experiments, based on a carefully designed and rigorous user study, demonstrate that, compared to closed-source state-of-the-art video AIGC models such as Pika, Kling, and Gen-3, our framework provides significantly more flexible control over object motions, generates videos that better align with user intentions, and achieves superior physical realism—all while maintaining comparable rendering quality.

\begin{figure*}[t]
\centering
\vspace{-15pt}
\includegraphics[width=\textwidth]{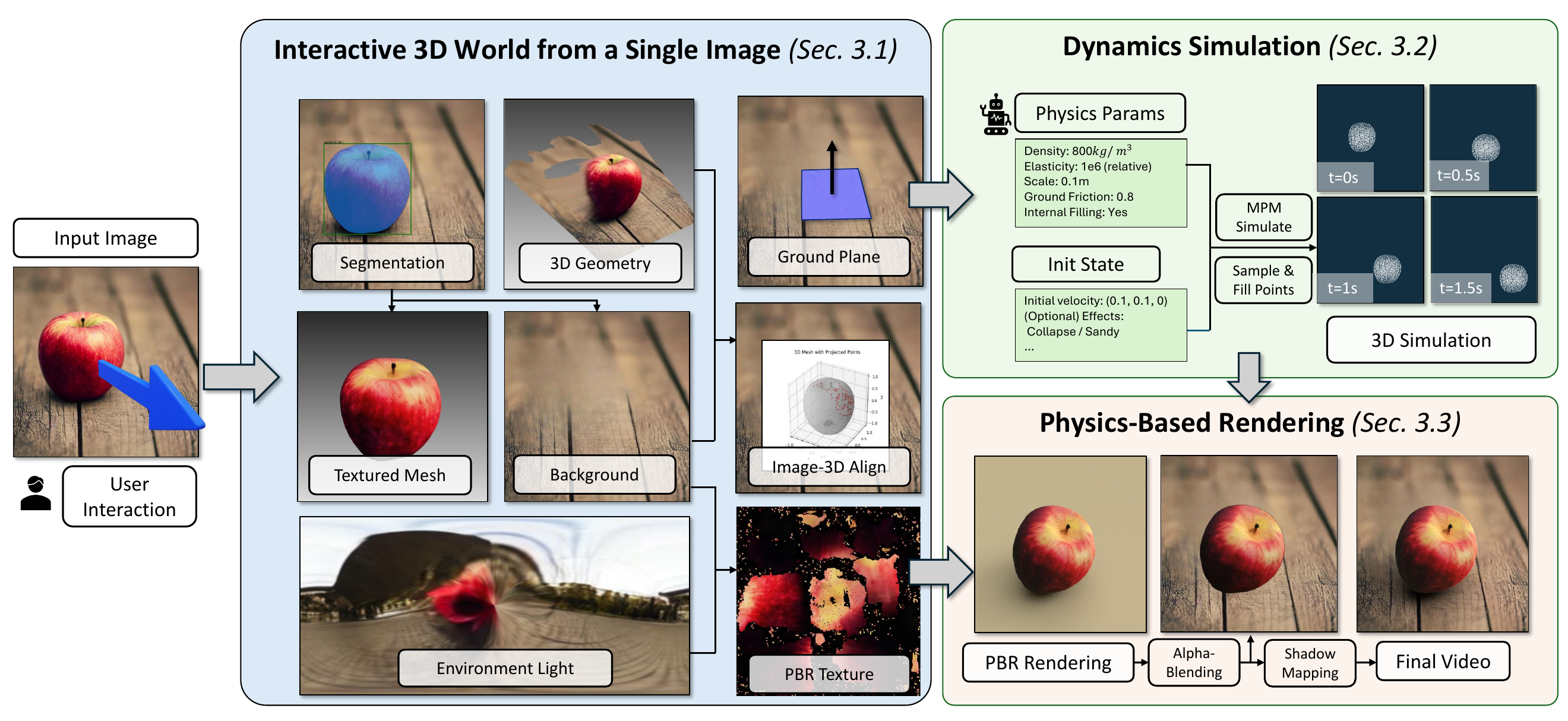}
\vspace{-10pt}
\caption{\textbf{Method Overview.} PhysGen3D's framework consists of three modules: a) 3D world creation, which infers geometry, semantics, rendering and physical parameters from the input image; b) dynamics simulation using Taichi-Elements for particle-based physics; and c) physics-based rendering with a two-pass shadow mapping technique.
}
\label{fig:method}
\vspace{-10pt}
\end{figure*}

\section{Related work}
\label{sec:related}

\paragraph{\textbf{Single-view 3D reconstruction}}
Great progress has been made in object-centric single view 3D reconstructions~\cite{liu2023zero1to3,shi2023zero123plus,xu2024instantmesh,liu2024one}. However, applying these techniques to single-view 3D scene reconstruction becomes more challenging. Most existing works only focus on one part of 3D scene understanding. Geometric approaches reconstruct the 3D scene holistically but neglect individual object understanding, or they focus solely on foreground objects without considering the interaction of objects with the complex environment. Additionally, works have been done for scene relighting~\cite{zhang2022simbar,liu2023deep, zheng2023steps} and segmentation~\cite{kirillov2023segany,ren2024grounded,liu2023grounding, yan20203d, zhong2020seeing}, but none of these provide a full 3D understanding within the image. In our work, we also perform perception reasoning for object materials, backgrounds, rendering properties, and physical stability.

\vspace{-5pt}
\paragraph{\textbf{Controllable video generation.}}
Video generation has made significant progress in recent years~\cite{bar2024lumiere,sora,blattmann2023stable,blattmann2023videoldm,ge2023preserve,girdhar2023emu,ho2022imagen,kondratyuk2023videopoet,gupta2023photorealistic,yu2023language,villegas2022phenaki,singer2022make,wang2023lavie,chen2023seine,xing2023dynamicrafter,2023I2VGen-XL}. The state-of-the-art framework~\cite{Pika2024website,klingai2024website} can generate photorealistic and coherent videos from text instructions using diffusion models~\cite{sohl2015deep,song2019generative,ho2020denoising,rombach2022high,Peebles2022DiT}. 
Controllable video generation is achieved through leveraging pre-trained video generation models with conditioning information, including but not limited to depth maps~\cite{chen2023control,yan2023magicprop}, linear translation~\cite{khachatryan2023text2video}, layouts~\cite{lian2023llm,wang2024boximator,yang2023diffusion}, and multiple combinations~\cite{ma2023trailblazer}. However, most existing generation methods 
implicitly generate image space dynamics,
which might lead to unrealistic, hallucinated motions.
In contrast, our method explicitly controls the motion and interaction with simulation, allowing us to create more sophisticated effects without the need for extensive training data. Our approach is training-free and generalizable to all objects in the world.


\vspace{-5pt}
\paragraph{\textbf{Controllable image animation}}
Image-based animation aims to animate objects that appear in images. Numerous works~\cite{chuang2005animating, davis2015image, xue2018visual, holynski2021animating, schodl2000video, Szummer96a, wei2000fast} have focused on this task. To improve quality, recent research has adopted data-driven solutions, training temporal neural networks to directly predict subsequent video frames~\cite{holynski2021animating, Blattmann_2021_CVPR, endo2019animating, chen2023livephoto, 2023VideoComposer, guo2023animatediff}, or incorporating physical heuristics~\cite{jhou2015animating, chuang2005animating}. Recently, there has been increased focus on interactive~\cite{blattmann2021ipoke, li2023generative, blattmann2021understanding} and controllable~\cite{davis2015image, aberman2020unpaired} image-to-video synthesis. Additional priors such as motion fields~\cite{endo2019animating, holynski2021animating, mahapatra2022controllable, mallya2022implicit, sugimoto2022water, xue2018visual}, optical flow~\cite{geng2024motion, zhao2022thin, bowen2022dimensions}, 3D geometry information~\cite{weng2019photo, qiu2023relitalk}, and user annotations~\cite{li2023generative} have also been introduced.
The closest work to ours is \citet{liu2025physgen}, which focuses on image understanding and uses an existing 2D rigid simulator to generate 2D animations. However, their approach is limited to 2D dynamics, does not account for real-world 3D physics, and is restricted to rigid bodies. In contrast, our method provides more realistic and flexible control over 3D motions and object materials and extends beyond rigid-body physics.

\section{Approach}

Our goal is to reconstruct an interactive, camera-centric miniature world from a single input image. We aim to control object materials, dynamics, and motions to simulate diverse, photorealistic, and physically plausible videos from user input. The key challenge lies in partial observations and the ill-posed nature of physical reasoning without observed dynamics. To address this, we propose a holistic reconstruction method leveraging large pretrained visual models to jointly infer geometry, dynamic materials, lighting, and PBR materials from a single image (\cref{sec:asset}). The reconstructed scene is input into a material-point-method (MPM) simulator to generate realistic physics phenomena (\cref{sec:sim}). Finally, we render dynamic object behaviors based on the simulation and reintegrate them into the scene, producing realistic videos with accurate appearance and motion (\cref{sec:render}). Fig.~\ref{fig:method} depicts our framework.



\subsection{Interactive 3D World from a Single Image}
\label{sec:asset}
A full amodal reconstruction of the 3D world depicted in the image is critical for the next step in simulation. An ideal reconstruction should capture a comprehensive understanding of objects' relationships, geometry, appearance, material, and physical properties. However, obtaining this understanding from a single image is highly ill-posed. Our key idea is to leverage priors from pretrained vision foundation models to help infer these properties, as shown in \cref{fig:method}.

\textbf{\textit{Segmentation}}. 
We leverage vision foundation models to recognize object categories and segment object instances. Specifically, we use GPT-4o \cite{yang2023dawn} to identify the foreground object categories and use Grounded-SAM \cite{kirillov2023segany,liu2023grounding,ren2024grounded} to further detect and segment each individual instance $\{o^i\in\mathbb{R}^{W\times H\times 3}\}_i^N$, where $o^i$ is the image of i-th object.

\textbf{\textit{Mesh Generation}}. 
Unlike previous work \cite{liu2025physgen}, which used only 2D rigid-body physics, we extend to general-purpose 3D simulation. This requires a complete 3D representation of foreground objects. We adopt InstantMesh~\cite{xu2024instantmesh}, which uses Zero123++ \cite{shi2023zero123plus, liu2023zero1to3} to synthesize multi-view images from the segmented object image \(o^i\) and uses these images to reconstruct the 3D mesh \(\mathcal{O}\) of the objects. For multiple object occlusions, we adopt the iterative inpainting strategy to extract the 3D mesh of the objects (see supp for more details).

\textbf{\textit{Background Handling}}.
The background serves important roles in both dynamic simulation and rendering. In simulation, background geometry acts as a support and collider, where accuracy ensures realistic object-scene interactions.
In rendering, it serves as a static backdrop while the foreground objects move and helps simulate realistic global illumination effects like cast shadows. For simulation purposes, we use Dust3r \cite{wang2024dust3r} to estimate iamge depth. The output depth map ${z \in \mathbb{R}^{W \times H}}$ is unprojected to the 3D world as a 3D point cloud $\mathcal{P}$, and Bilateral Normal Integration~\cite{cao2022bilateral} is applied to generate a smooth surface $\mathcal{S}$ serving as the collider. For rendering, to fill background regions occluded by moving objects, we use the LaMA inpainting model\cite{suvorov2021resolution}. This model generates a complete background after masking out all objects and their shadows.

\textbf{\textit{Object Pose and Scale Estimation}}. 
The generative 3D reconstruction step provides a complete mesh with a normalized scale in an object-centric coordinate system, but it does not infer the object's location and scale in the camera coordinate.  
To ensure coherence with the input image, it is necessary to accurately place the 3D meshes $\mathcal{O}$ into the 3D scene $\mathcal{P}$ with the correct scale and 6DoF pose. 
This registration task is challenging, as the generated object mesh may not perfectly match the real-world object in the image. 

To address this challenge, we designed a multi-stage coarse-to-fine alignment strategy. 
In the coarse stage, we perform 2D-3D feature point matching.
Firstly, we render multiple images from viewpoints uniformly distributed over a unit sphere surrounding the object. For each rendered image, we match its feature points with the original object image $o^i$ using SuperGlue \cite{sarlin2020superglue}, and the viewpoint with the most matches is selected. Matched pairs $({p'}_i^N \in \mathbb{R}^2, p_i^N \in \mathbb{R}^2)$ are then projected back into 3D points $({P'}_i^N \in \mathbb{R}^3, P_i^N \in \mathbb{R}^3)$ in object and camera coordinate, respectively.
Perspective-n-Point (PnP) algorithm~\cite{fischler1981random} is applied between ${P'}_i^N$ and ${p}_i^N$ in image $o^i$ to estimate the 6DoF pose with scale ambiguity. Then we adjust the scale and translation factor simultaneously to minimize the L2 loss \(\sum_i^N = ||P_i^N - {P'}_i^N||_2\), without altering its projection.

In the fine alignment stage, we render the mask and depth on the image plane using the current estimation through a differentiable renderer and jointly minimize two losses: $\mathcal{L} = \mathcal{L}_\mathrm{dice} + \mathcal{L}_\mathrm{depth}$.
Here, $\mathcal{L}_\mathrm{dice} = 1 - \frac{2 \times |M_A \cap M_B|}{|M_A| + |M_B|}$ measures the discrepancy between the rendered mask $M_A$ and the observed mask $M_B$ from Grounded-SAM.
The depth consistency loss is $\mathcal{L}_\mathrm{depth} = \frac{\mathrm{mse}(M_B \ast z_A - M_B \ast z_B)}{|M_B|}$, where $z_A$ and $z_B$ are the rendered and Dust3r-predicted depths. 
This joint optimization ensures consistency between the estimated mesh in the camera coordinate and the point cloud from Dust3r, while maximizing the alignment between the 3D pose and observed object mask, ensuring accurate simulation and rendering alignment.

\textbf{\textit{Appearance Optimization}}. The texture of the generated 3D meshes may differ from the input image. To enhance rendering quality, we use the inverse rendering pipeline in Mitsuba3 \cite{Jakob2020DrJit} to estimate material properties. Lighting parameters are estimated with DiffusionLight \cite{phongthawee2024diffusionlight}, while object PBR materials (albedo, roughness, and metallic) are optimized via differentiable rendering. To handle unknown back views and reduce complexity, we assume uniform roughness and metallic values per object and apply tone mapping for albedo optimization:
\(y(x) = ax^3 + bx^2 + cx + d, \quad \text{where } y(0) = 0, \, y(1) = 1.\)
With reconstructed lighting and ground surface, the optimized materials improve asset appearance, capturing realistic object-surface interactions during rendering.

\textbf{\textit{Physics Reasoning}}.
Accurately simulating real-world dynamics requires estimating physical parameters. We focus on two aspects: 1) Following \cite{xia2024video2game}, we use GPT-4o to query each object's elasticity and density, and the friction coefficient for the surface \(\mathcal{S}\). 2) We ensure reconstructed 3D assets match real-world proportions, as depth inaccuracies from Dust3r can cause unrealistic behaviors. To address this, we estimate a scale factor \(k\) by comparing asset size with typical real-world sizes from GPT-4o and use \(k\) for dimensionless scaling of gravity and velocity-related parameters.

\begin{figure*}[t]
    \centering
    \vspace{-13pt}
    \includegraphics[width=0.8\linewidth]{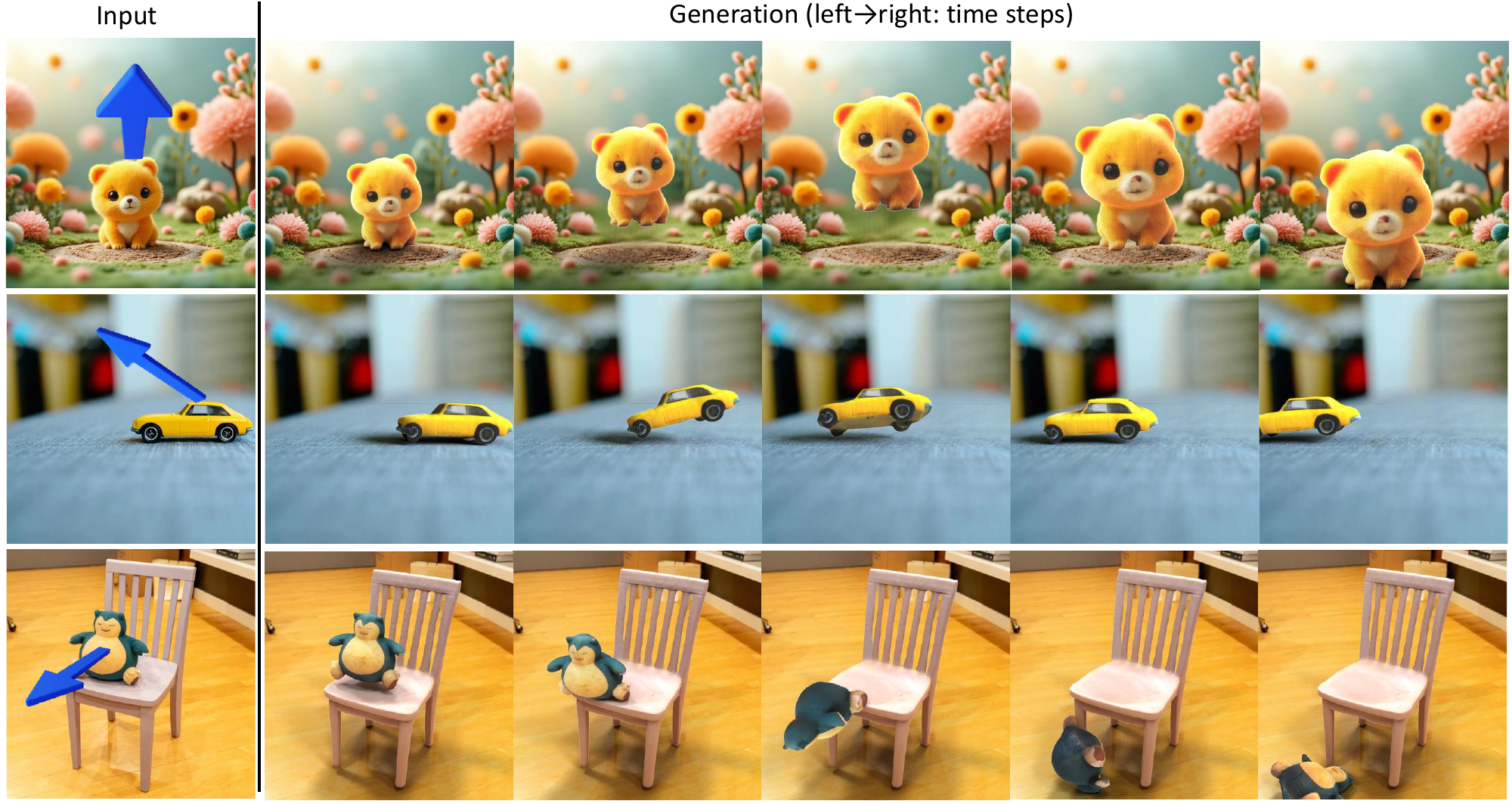}
    \vspace{-5pt}
    \caption{\small
    \textbf{Video generation results.}
     Left: Input initial frame. Right: Generated future frames. We apply an initial velocity to each movable object and use the physically grounded parameters outlined in \ref{physicalpara} to generate physically plausible results.
    }
    \label{fig:qual}
    \vspace{-13pt}
\end{figure*}
\subsection{Dynamics Simulation}
\label{sec:sim}
Given the 3D assets with reasoned physical properties and the scale factor, we use the physics engine Taichi Elements \cite{hu2019difftaichi, hu2019taichi, hu2021quantaichi}, based on the Material Point Method (MPM) \cite{jiang2016material}, as our simulator to support a variety of different materials, including but not limited to rigid, soft, and granular. 

\textbf{\textit{Particle representation}}. The simulator is based on a particle-based representation. To convert our 3D assets into a simulatable particle-based representation, we apply floating points removing, internal filling and voxel downsampling. We use downsampling to handle uneven particle distributions, where downsample rate is adjusted according to the grid size of the simulator. For the convenience of rendering, we prioritize surface points. 

\textbf{\textit{Physical parameters}}.\label{physicalpara} To enhance stability, we apply the scale factor \(k\) to physical parameters in simulator instead of scaling the assets to real size. For example, the gravitational acceleration is multiplied by \(k\), making to \(k*9.8\) (nondimensionalized here). With this tuning, the motion of falling or collapse remains realistic for all scales.

\textbf{\textit{External disturbance}}. For each object, we set a different initial velocity based on the user input to make the object move as specified by the user.

\textbf{\textit{Other Visual Effects}}. Besides real world physics simulation, our pipeline allows special effects like collapsing and melting. To simulate different materials (rigid, soft, or granular), we can easily change the material type to modify the physical properties of an object. This approach provides more flexibility for the user to edit the scene.


\subsection{Physics-Based Rendering}
\label{sec:render}
After dynamics simulation, we obtain object point trajectories and apply motion interpolation to compute vertex motions, deforming the mesh accordingly. With optimized PBR materials, we use Mitsuba3 for Physically-Based Rendering under environment lighting. Following prior insertion rendering work~\cite{karsch2011rendering, geosim, chatsim, unisim}, we avoid converting the entire static background into the rendering pipeline. Instead, we build a 3D shadow catcher surface from the background depth. During rendering, no texture is applied to the background; two-pass shadow mapping extracts shadows and global illumination effects. The foreground objects and shadows are then composited onto the inpainted background to produce the final video with realistic lighting.

\begin{figure*}[t]
    \centering
    \vspace{-13pt}
    \includegraphics[width=0.8\linewidth]{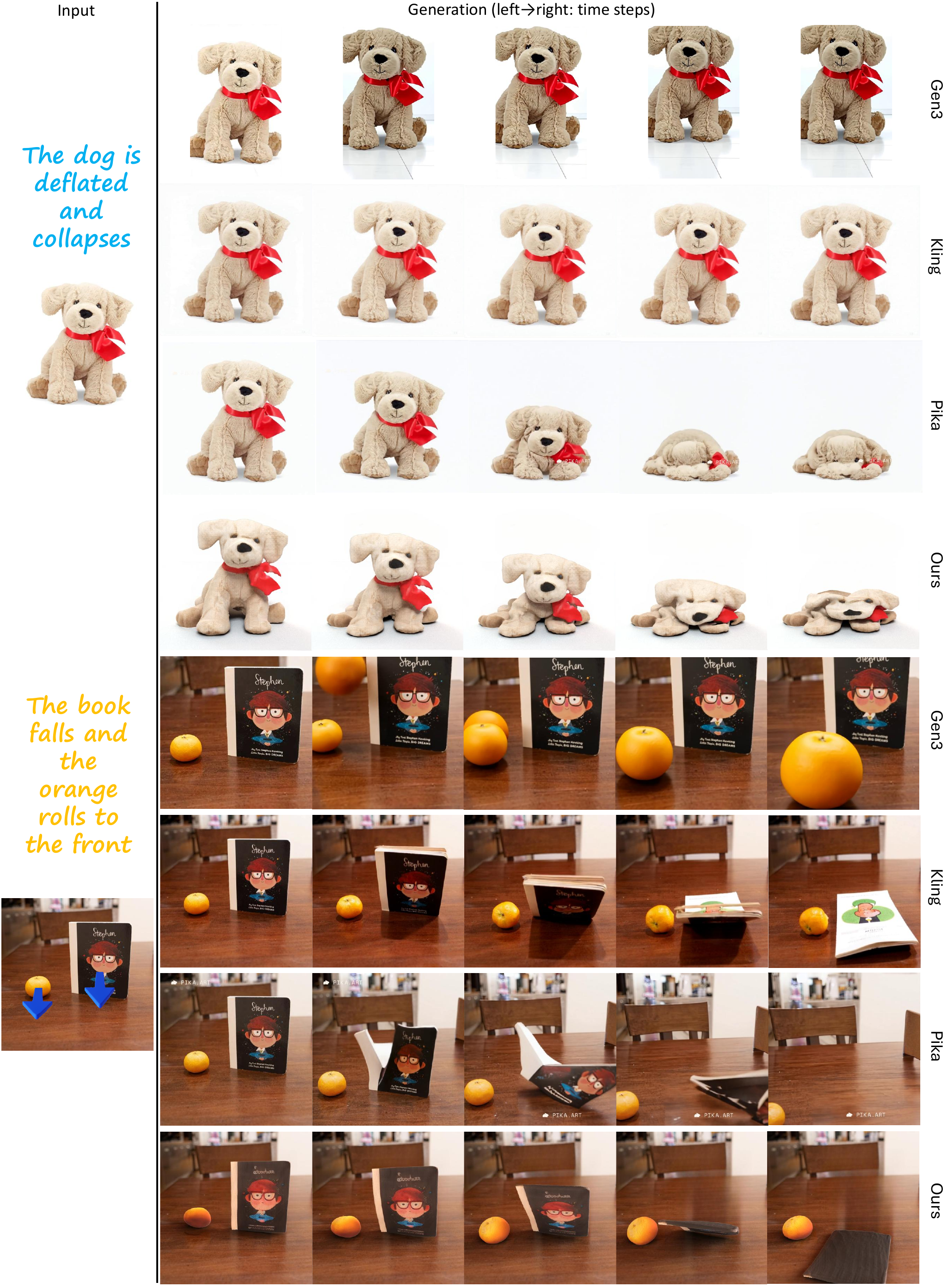}
    \caption{
    \textbf{Qualitative comparison.}
    We compare videos generated from our framework with three state-of-the-art I2V models: Gen-3 \cite{Runway2024website}, Pika \cite{Pika2024website} and Kling \cite{klingai2024website}. We carefully designed the prompt to describe the motion outcome, and uses motion brush to control Kling. Our framework employs initial velocity control. Results show that our method can follow text instructions while maintaining plausible physics.
    }
    \label{fig:comparison}
    \vspace{-13pt}
\end{figure*}
\begin{figure*}[t]
    \centering
    \vspace{-13pt}
    \includegraphics[width=\linewidth]{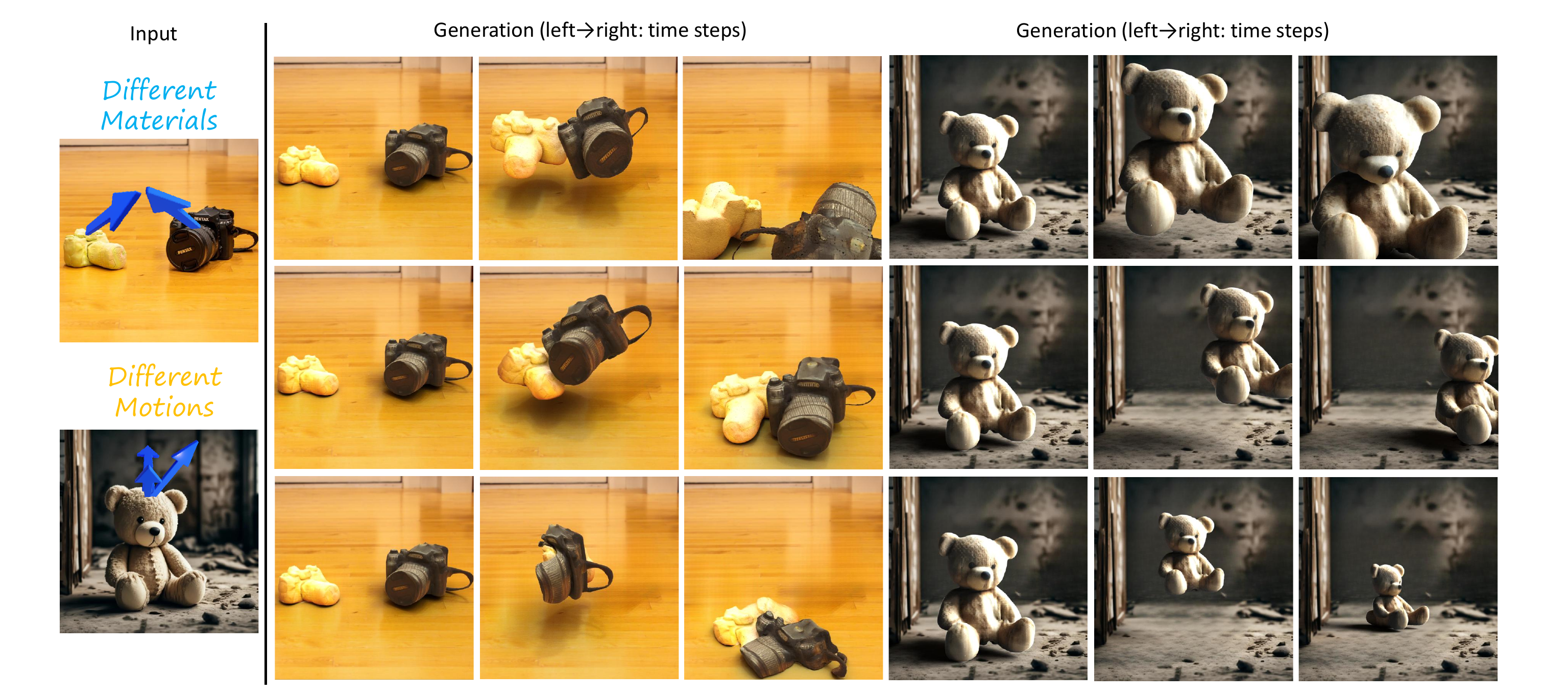}
    \caption{\small
    \textbf{Dynamics Effects.} We can generate various dynamics from the same input image. The left three columns share the same initial positions and velocities, but are in different materials. The right three columns have the same material, but defer in velocity directions. This showcases the potential of our method for generating diverse physical scenarios.
    }
    \vspace{-5pt}
    \label{fig:effects}
\end{figure*}
\begin{figure*}[t]
    \centering
    \includegraphics[width=0.8\linewidth]{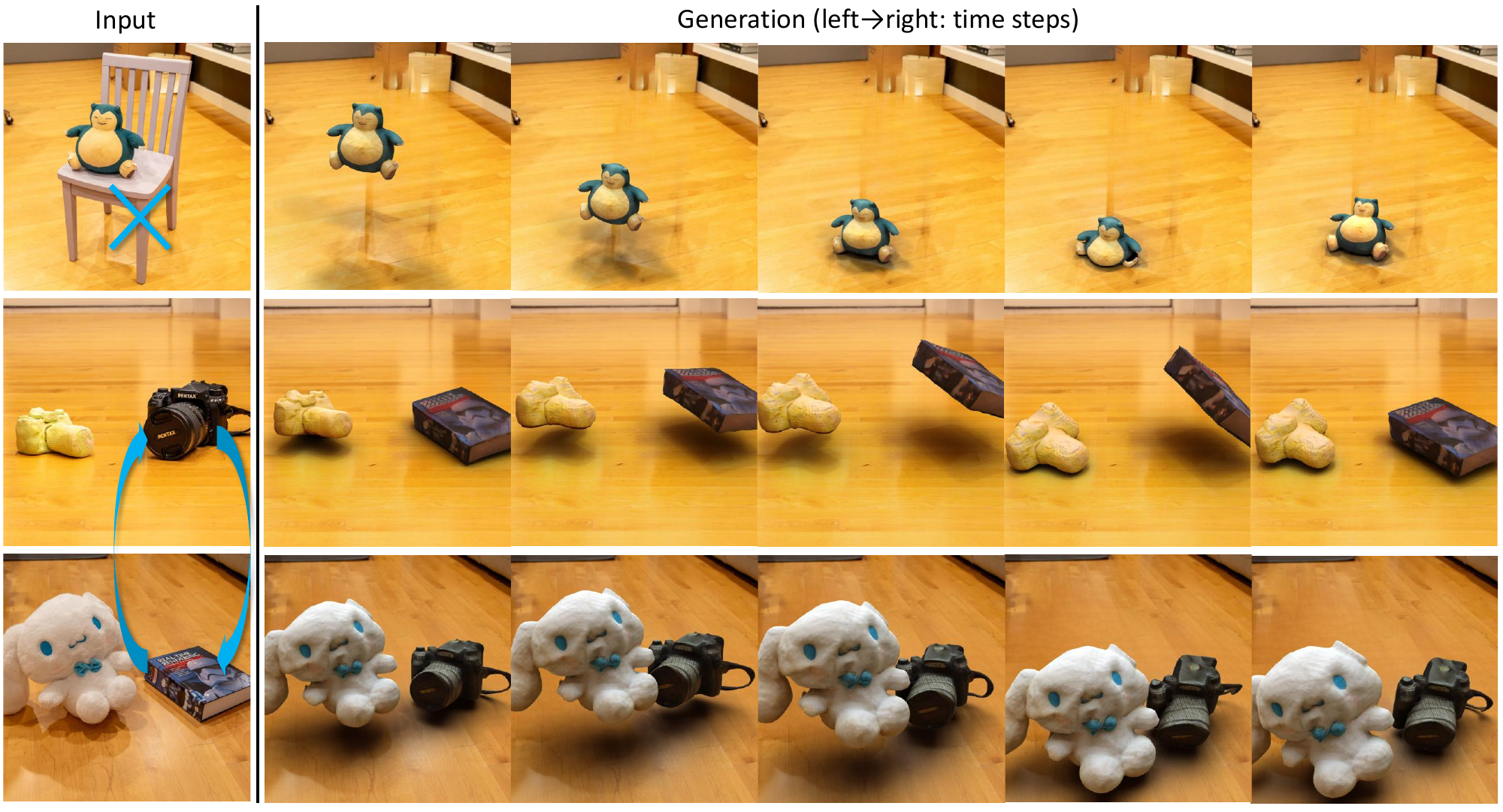}
    \caption{\small
    \textbf{Video edition.} As illustrated, our method supports video editing. In the top row, we remove the chair and allow the toy to fall from a static position. In the bottom two rows, we exchange one object between two scenes while keeping the other unchanged. This demonstrates the great flexibility of our video generation approach.
    }
    \label{fig:edit}
    \vspace{-8pt}
\end{figure*}

\section{Experimental Results}
\subsection{Setups}
The test images come from a diverse set of our own photography, online collections, and generative models. Our pipeline is primarily designed for object-centric scenes with one or a few objects. Due to our limitations (Sec.~\ref{limitation}), we excluded those with excessive objects, heavy occlusion between objects, or highly uneven surfaces.

\textbf{\textit{Post-processing}}.
We used VEnhancer ~\cite{he2024venhancer} as an optional post-processing module, which takes the produced video and a text prompt to perform enhancement. As shown in \Cref{tab:quantitative}, while it restores some details, it also introduces extra hallucinations. 

\textbf{\textit{Baselines}}. 
Our method is one of the first of its kind, as existing model-simulate approaches require multi-view images~\cite{video2game, zhang2025physdreamer, xie2023physgaussian} or specific settings~\cite{liu2025physgen, li2023generative, sugimoto2022water}. Therefore, we evaluate ours against Image-to-Video (I2V) models: two open-source motion controller models DragAnything\cite{wu2024draganything}, MOFA-Video\cite{niu2024mofa} and three state-of-the-art (SOTA) commercial models Kling 1.0 \cite{klingai2024website}, Gen-3 \cite{Runway2024website}, and Pika 1.5 \cite{Pika2024website}.
We manually set correct motion trajectories and select applied regions for DragAnything, MOFA-Video and Kling to provide privileged motion guidance. We give text descriptions to Pika 1.5 and Gen-3, as they lack direct motion control capabilities. Additionally, Pika 1.5 offers "Pikaffect" effects, such as "Melt it" and "Deflate it."


\subsection{Results}
Our system generates a miniature interactive world from a single image, enabling the simulation of various phenomena. \Cref{fig:qual} presents videos generated from different types of images. These images encompass single or multiple objects and various physical materials (i.e., rigid or soft). We also show applications like dynamic changes, object editing, and dense 3D tracking, illustrating the adaptability and creative potential of our approach for generating customized, interactive video content.

\textbf{\textit{Comparisons}}. We compare the results in two dimensions: motion control alone and  physical materials. \Cref{fig:comparison} shows that our system produces more physically plausible and controllable videos compared to SOTA I2V models. Despite prompt tuning, learning-based models often hallucinate, failing to adhere to physical laws or user intent. For example, in the toy dog case, we manually adjust the material and accurately simulate a collapse, whereas other models fall short. Similarly, for the book case, our results are the most physically realistic.

\textbf{\textit{Dynamics}}. \Cref{fig:effects} demonstrates the varying dynamics generated from the same input image, highlighting the high controllability of our method over physical parameters and motion trajectories. In the three rows on the left, we set different elasticities for the two objects, while keeping their initial positions and velocities the same. In the three rows on the right, we alter velocity directions of the objects and keep physical parameters the same.

\textbf{\textit{Editing}}. Our method enables modifications to videos by removing, adding, or replacing objects, as illustrated in \Cref{fig:edit}. The generated 3D assets can be easily manipulated, allowing for diverse video edits.

\begin{figure}[t]
    \vspace{-16pt}
    \centering
    \includegraphics[width=0.8\columnwidth]{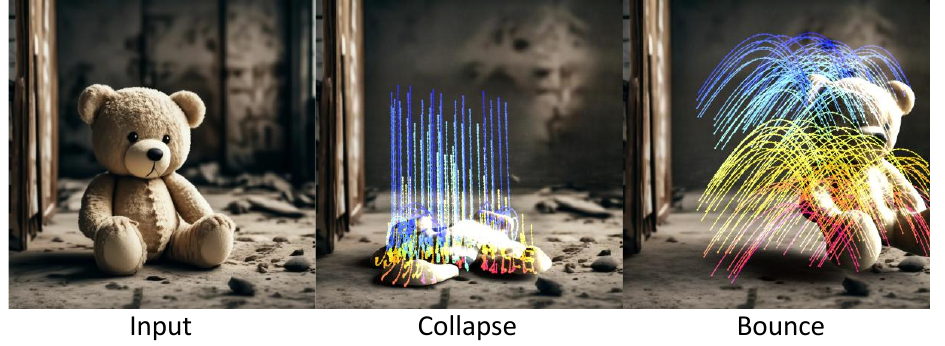}
    \vspace{-5pt}
    \caption{\small
    \textbf{Dense 3D Tracking.} Our method can naturally generate videos with dense 3D tracking results. Here we show tracking results for collapse and bounce cases.
    }
    \vspace{-5pt}
    \label{fig:tracking}
\end{figure}

\textbf{\textit{Tracking.}} Our framework uses an explicit 3D representation and works with a particle-based physics simulator. This allows our method to easily create videos with detailed 3D tracking results. \Cref{fig:tracking} showcases two examples, demonstrating the accuracy and reliability of the tracking in different scenarios.

\begin{figure}[t]
    \centering
    \includegraphics[width=\columnwidth]{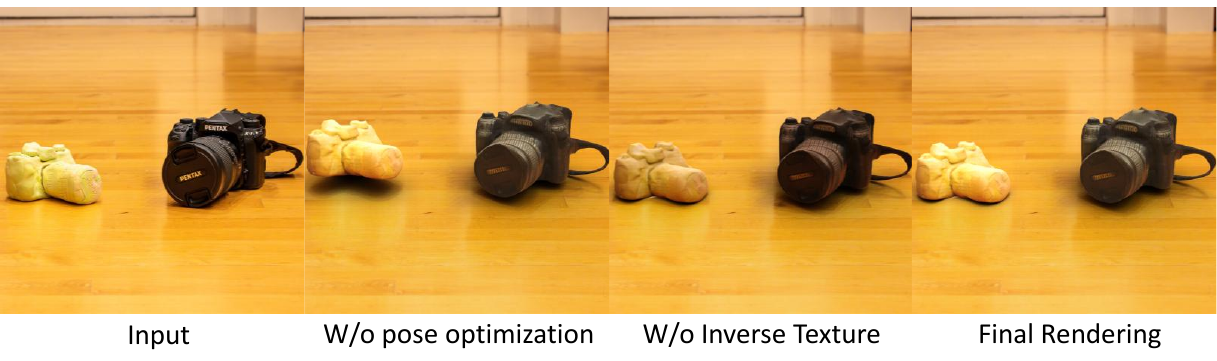}
    \vspace{-15pt}
    \caption{\small
    \textbf{Ablation study.} 
    Without pose optimization, the two objects cannot fully replicate the scene. Without inverse texture, the object mesh does not match the input color. 
    }
    \vspace{-5pt}
    \label{fig:ablation}
\end{figure}

\textbf{\textit{Ablation}}. Each step we designed in perception, simulation and rendering is intended to mimic the real world. \Cref{fig:ablation} shows the results of ablation study involving position optimization and inverse texture. Without pose optimization using differentiable rendering, the two objects are roughly at the initial place shown in input image, but cannot fully replicate the scene. Without inverse texture, the generated object mesh does not match the input in terms of color tone and brightness. We also conducted ablations on point sampling. However, without sampling, a large number of points become crowded in several MPM simulation grids, causing the simulator to crash and fail to produce final outputs.


\begin{table}[htbp]
    \centering
    \caption{\textbf{Human Evaluation Results.} The three criteria are: Physical Realism, Photorealism and Semantic Consistency}
    \vspace{-5pt}
    \resizebox{.7\columnwidth}{!}{
    \begin{tabular}{ccccc}
        \toprule
        \textbf{Methods} & \textbf{PhysReal} & \textbf{PhotoReal} & \textbf{Align} \\
        \midrule
        Kling 1.0 \cite{klingai2024website} & 2.811 & 3.566 & 2.467 \\
        Runway Gen-3 \cite{Runway2024website} & 2.283 & \textbf{3.582} & 1.886 \\
        Pika 1.5 \cite{Pika2024website} & 2.412 & 3.314 & 2.016 \\
        Ours & \textbf{3.707} & 3.411 & \textbf{3.866} \\
        \bottomrule
    \end{tabular}
    }
    \label{tab:human_eval}
    \vspace{-5pt}
\end{table}


\begin{table}[htbp]
    \centering
    \caption{\textbf{VBench Scores and GPT-4o Evaluation Results.} \textbf{Motion} and \textbf{Imaging} refers to Motion Smoothness and Imaging Quality scores in VBench. The three criteria on the right are the same as \Cref{tab:human_eval}, with results given by GPT-4o.}
    \vspace{-5pt}
    \resizebox{\columnwidth}{!}{
    \begin{tabular}{c|cccccc}
        \toprule
        Methods    & \textbf{Motion$\uparrow$} & \textbf{Imaging$\uparrow$} & \textbf{PhysReal$\uparrow$}  & \textbf{PhotoReal$\uparrow$} & \textbf{Align$\uparrow$}\\
        \midrule
        Kling 1.0            & \textbf{0.996} & 0.671 &	0.563 & 0.874 & 0.596\\
        Runway Gen-3        &	0.991& \textbf{0.723} &0.141 & \textbf{0.896} & 0.144 \\
        Pika 1.5           &  0.994 & 0.671  & 0.544 & 0.863 & 0.563  \\
        MOFA-Video          &  0.994 & 0.634 & 0.384  &   0.764   & 0.304  \\
        DragAnything    & 0.985 & 0.428	& 0.645 & 0.756   &	0.380 \\
        \textbf{Ours}   & \underline{0.995} & 0.666 & \underline{0.752} & 0.867 & \textbf{0.796}\\
        \textbf{Ours+VEnhancer}   & 0.994 & \underline{0.677}& \textbf{0.766} & \underline{0.880} & \underline{0.745}\\
        \bottomrule
    \end{tabular}
    }
    \label{tab:quantitative}
    \vspace{-10pt}
\end{table}

\subsection{Quantitative Comparison}

To assess the quality of the generated videos, we performed human evaluation, GPT-4o evaluation and provided two VBench\cite{huang2024vbench} evaluation scores.

\textbf{Benchmarks.}
We designed three criteria for human and GPT-4o evaluation. (1)\textbf{Physical Realism (PhysReal)} measures how realistically the video follows the physical rules and whether the video represents real physical properties like elasticity and friction. (2) \textbf{Photorealism (Photoreal)} assesses the overall visual quality of the video, including the visual artifacts, discontinuities, and how accurately the video replicates details of light, shadow, texture, and materials. (3) \textbf{Semantic Consistency (Align)} evaluates how well the content of the generated video aligns with the intended meaning of the text prompt.
We also chose two Quality Scores in VBench: Motion Smoothness and Imaging Quality.

\textbf{\textit{Details in evaluation}}.
Following a similar methodology to \cite{liu2025physgen}, we designed a questionnaire with 27 videos covering various scenes, motion conditions, and effects. Each video includes an input image, a motion prompt, and outputs from three SOTA commercial baselines and ours, shown in random order. 31 participants rated three criteria on a five-point scale from strongly disagree (1) to strongly agree (5). We also evaluated all five baselines and our diffusion-enhanced version using GPT-4o and VBench. GPT-4o assessed videos on the same criteria based on 10 evenly sampled frames, with the input image and prompt provided. We used GPT version gpt-4o-2024-08-06, with detailed prompts in the supp.

\textbf{\textit{Result analysis}}.
The results in \Cref{tab:human_eval} demonstrate our ability to generate physically accurate and controllable videos. In physical realism (\textbf{PhysReal}) and semantic consistency (\textbf{Align}), our method achieves the highest scores and outperform all the commercial models by a large margin.  \Cref{tab:quantitative} presents the results of GPT-4o and VBench evaluation. GPT score is aligned with human evaluation, and we also beat open-sourced in VBench. Among the baseline models, Kling 1.0 performs the best overall, likely due to its use of the "motion brush", which specifies motion trajectories.





\section{Limitations}
\label{limitation} 

\begin{figure}[t]
    \vspace{-16pt}
    \centering
    \includegraphics[width=\columnwidth]{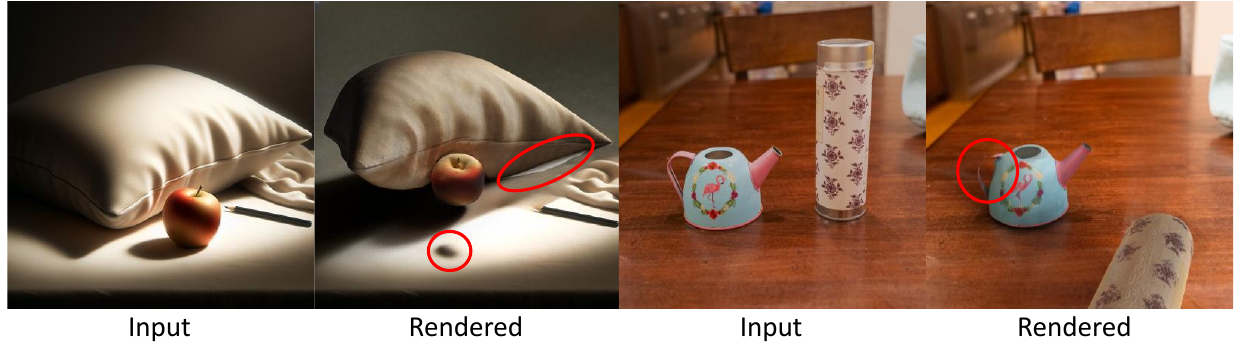}
    \vspace{-13pt}
    \caption{\small
    \textbf{Limitations.}
    The first set of two images show rendering errors: extreme lighting and heavy shading cause incorrect painting and artifacts from the pillow penetrating the ground. The second set reveals simulation limitations: the MPM method struggles with fine details, such as the thin kettle handle, leading to failures.
    }
    \label{fig:limit}
    \vspace{-10pt}
\end{figure}

Our single-image-based interactive miniature world is designed for object-centric scenes with simple spatial geometry and inter-object relationships. Reconstructing complete scenes for more complex scenarios remains an open challenge. 
\Cref{fig:limit} illustrates several failure cases, such as rendering errors under challenging shading, perception failures, and simulation limitations. More detailed analysis on limitations can be found in the supp.
\section{Conclusions}
We present PhysGen3D, a framework that transforms a static image into an interactive 3D scene for simulating and rendering future motions based on user input. PhysGen3D integrates modules for 3D world reconstruction, model-based dynamic simulation, and physics-based rendering to generate realistic, controllable videos. By extending the 2D image-to-video paradigm to 3D, PhysGen3D enables more realistic motion and diverse material behaviors. We hope this work inspires future research.

\clearpage
\section*{Acknowledgement}
{
\small
This project is supported by the Intel AI SRS gift, Amazon-Illinois AICE grant, Meta Research Grant, IBM IIDAI Grant, and NSF Awards \#2331878, \#2340254, \#2312102, \#2414227, and \#2404385. We greatly appreciate the NCSA for providing computing resources.
}
{
    \small
    \bibliographystyle{ieeenat_fullname}
    \bibliography{main}
}

\appendix
\maketitlesupplementary

In the supplemental materials, we present additional details about our PhysGen3D framework \cref{sec::add_method}, more details of our experimental design \cref{sec::design}, more quantitative and qualitative results \cref{sec::add_results}, and various applications of our system \cref{sec::app}. Furthermore, we invite the reviewers to check a local webpage in the supplemental materials accessed by \texttt{index.html}, to see our generated videos.

\section{Additional Details of PhysGen3D}
\label{sec::add_method}

We provide additional details about our framework, specifically on how we handle multiple object occlusions during the mesh generation stage, how we address background completion concerning objects and their shadows, the detailed prompt used in physics reasoning, and further specifics about the physical simulator utilized in our approach.

\begin{figure}[b]
    \centering
    \includegraphics[width=\columnwidth]{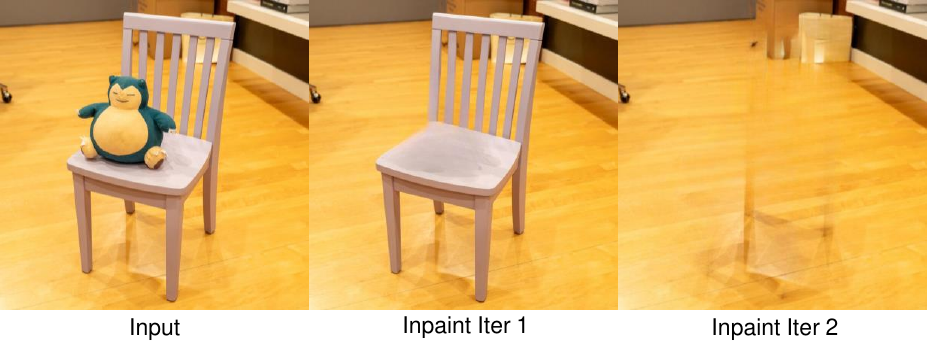}
    \caption{\small
    \textbf{Iterative Inpainting.} Left: Input image. Middle: Inpainting result after 1 iteration, where the toy is masked and inpainted. Right: Inpainting result after 2 iterations, where the chair is masked and inpainted. The second result is used as background. 
    }
    \label{fig:mesh_generation}
    \vspace{-5pt}
\end{figure}
\subsection{Mesh Generation}
To reconstruct a 3D foreground object, we require a complete and clearly segmented object image \(o^i\). For scenarios with multiple object occlusions, we employ an iterative inpainting and segmentation strategy, as illustrated in \Cref{fig:mesh_generation}. We first identify all the target objects using GPT-4o. In cases where occlusions are detected, the objects are segmented and inpainted sequentially, progressing from the foreground to the background. Each subsequent segmentation step builds upon the removal of previously processed objects, ensuring accurate and unobstructed reconstruction.

\subsection{Background Handling}
Shadow significantly impacts the quality of background inpainting if not masked properly. Existing shadow removal methods \cite{li2024shadow, liu2021shadow, vasluianu2024ntire} typically detect and remove all shadows indiscriminately. However, our goal is to remove only the shadow related to a specific object. To achieve this, we adopt a straightforward method: we first segment regions where brightness values fall below a certain threshold to identify shadows. For each object, we determine the largest connected component that includes both the object and its shadow. Then, we dilate this mask with a kernel of size 50 and apply inpainting. Developing more adaptable, per-object shadow removal techniques is left as future work.

\subsection{Physics Reasoning}
\lstset{
  basicstyle=\ttfamily\footnotesize,
  backgroundcolor=\color{gray!10},
  frame=single,
  breaklines=true,
  breakatwhitespace=true,
  columns=flexible,
}
We use GPT-4o to reason the physical parameters for each object and the surface. The prompt and an example answer are as follows.
\begin{lstlisting}[caption={Prompt used for GPT-4o physics reasoning}]
Answer each question for each object in the picture, using one word or number, separated by commas. For numbers, do not use scientific notation.
Provide answers in the following format for each object:
'Object number, name, density in kg/m^3, Young's modulus (soft/medium/hard), size in meters, requires internal filling (yes/no).'
If there are multiple objects in the picture, respond for each object on a new line in the specified format.
What is each object's name (one word)?
What is its density in kilograms per cubic meter?
What is its Young's modulus in Pa? (Choose from: Soft: Materials like plush toys, foam, or fabric. Medium: Materials like rubber or soft plastic. Hard: Materials like wood, metal, or hard plastic)
What is its size in meters?
Does the object require internal filling for MPM simulations (yes/no)?

Estimate the roughness of the supporting surface in the picture, such as tables, floors, or any other horizontal surfaces that can act as supports. Provide answers in roughness value (0 to 1, where 0 = perfectly smooth and 1 = extremely rough)'.
\end{lstlisting}

\begin{lstlisting}[caption={Example answers from GPT-4o}]
1, camera_model, 200, soft, 0.15, yes
2, camera, 2700, hard, 0.20, no
0.2
\end{lstlisting}

In our observation, GPT-4 often provides unstable results for the exact value of Young's modulus, with discrepancies spanning several orders of magnitude. To address this, we defined three categories—\textbf{soft}, \textbf{medium}, and \textbf{hard}—to guide GPT's classification. In the simulator, the elasticity \( E \) does not directly correspond to the real Young's modulus. Based on experience, we associate the three categories with \( E = 5 \times 10^4 \), \( E = 5 \times 10^5 \), and \( E = 5 \times 10^6 \), respectively.

\subsection{Dynamics Simulation}
For simulation stability, we fix the size of the simulator to 2 and the resolution to 256. Since the target object's scale varies from several centimeters to tens of meters, we align the object with the reconstructed scene and fit it into the simulator. To simulate real physics, we scale the physical parameters accordingly. Suppose the reasoned real size of the object is \( s_0 \), and the scaled mesh has size \( s' \). Then, the scaling factor is \( k = \frac{s'}{s_0} \). In the simulator, we set gravity to \( g' = k \times g_0 = k \times 9.8 \). The elasticity of each object is also scaled: \( E'_i = \frac{E_i}{k} \). (According to dimensional analysis, Young's modulus is inversely proportional to the scale of length.)

We use Taichi Elements \cite{hu2019difftaichi, hu2019taichi, hu2021quantaichi} for Material Point Method (MPM) simulations and modify it to support inhomogeneous materials. MPM is a computational technique used to simulate the behavior of continuum materials. The governing equation of motion is:
\[
\rho \frac{D\mathbf{v}}{Dt} = \nabla \cdot \boldsymbol{\sigma} + \mathbf{f}_{\text{ext}},
\]
where:
\begin{itemize}
    \item \(\rho\): Density of the material,
    \item \(\mathbf{v}\): Velocity field,
    \item \(\boldsymbol{\sigma}\): Cauchy stress tensor,
    \item \(\mathbf{f}_{\text{ext}}\): External forces per unit volume.
\end{itemize}

To be specific, MPM combines the strengths of Lagrangian and Eulerian methods by representing materials as discrete particles while performing computations on a background grid. The key steps of MPM are particle-to-grid (p2g) and grid-to-particle (g2p) transfers.

\paragraph{Particle-to-Grid (p2g) Transfer.}
This step transfers particle properties (mass, momentum, etc.) to the grid.

\textbf{\textit{Mass Transfer.}} Grid mass is computed by distributing particle mass \( m_p \) to nearby grid nodes using weighting functions \( w \):
\[
m_i = \sum_p w(x_p - x_i) m_p,
\]
where:
\begin{itemize}
    \item \( m_p = \rho_p V_p \): Particle mass (density \( \rho_p \), volume \( V_p \)),
    \item \( w \): Quadratic kernel for interpolation.
\end{itemize}

\textbf{\textit{Momentum Transfer.}} Momentum is transferred to the grid using the same weight:
\[
\mathbf{v}_i = \frac{\sum_p w(x_p - x_i) \mathbf{v}_p m_p}{m_i},
\]
where:
\begin{itemize}
    \item \( \mathbf{v}_i \): Grid velocity,
    \item \( \mathbf{v}_p \): Particle velocity.
\end{itemize}

\textbf{\textit{Stress Contribution.}} The stress tensor \( \boldsymbol{\sigma} \) contributes force to the grid momentum. Using the deformation gradient \( F \), the stress is defined as:
\[
\boldsymbol{\sigma} = 2 \mu (F - \mathbf{R}) F^\top + \lambda J (J - 1) \mathbf{I},
\]
where:
\begin{itemize}
    \item \(\mu\) and \(\lambda\): Lamé parameters,
    \item \(F\): Deformation gradient,
    \item \(\mathbf{R}\): Rotation matrix from SVD (\(F = \mathbf{R} \mathbf{S}\)),
    \item \(J = \det(F)\): Determinant of \(F\),
    \item \(\mathbf{I}\): Identity matrix.
\end{itemize}
The Lamé parameters \(\lambda\) and \(\mu\) are computed from Young’s modulus \(E\) and Poisson’s ratio \(\nu\) as follows:
\[
\lambda = \frac{E \nu}{(1 + \nu)(1 - 2\nu)}
\]
\[
\mu = \frac{E}{2(1 + \nu)}
\]
where:
\begin{itemize}
    \item \(E\): Young's modulus, which describes the material's stiffness,
    \item \(\nu\): Poisson’s ratio, which defines the ratio of lateral strain to axial strain.
\end{itemize}

\textbf{\textit{Grid Velocity Update.}} 
The grid force due to stress is given by:
\[
\mathbf{f}_i = - \sum_p w'(x_p - x_i) V_p \boldsymbol{\sigma}_p.
\]
Newton’s second law updates grid velocities:
\[
\mathbf{v}_i^{n+1} = \mathbf{v}_i^n + \Delta t \frac{\mathbf{f}_i}{m_i},
\]
where \( \Delta t \) is the time step.

\paragraph{Grid-to-Particle (g2p) Transfer}
This step interpolates updated grid data back to particles and updates their states (e.g., velocity, deformation).

\textbf{\textit{Velocity Interpolation.}} Particle velocities are updated by interpolating grid velocities:
\[
\mathbf{v}_p^{n+1} = \mathbf{v}_p^n + \sum_i w(x_p - x_i) \mathbf{v}_i^{n+1}.
\]

\textbf{\textit{Affine Velocity Field.}} Affine velocity updates capture velocity gradients from the grid:
\[
\mathbf{C}_p = \sum_i 4 \frac{w(x_p - x_i)}{\Delta x} \mathbf{v}_i \otimes (\mathbf{x}_i - \mathbf{x}_p).
\]

\textbf{\textit{Deformation Gradient Update.}} The deformation gradient \( F_p \) evolves based on the velocity gradient:
\[
F_p^{n+1} = (\mathbf{I} + \Delta t \mathbf{C}_p) F_p^n,
\]
where \( \mathbf{I} \) is the identity matrix.

\textbf{\textit{Advection.}} Finally, particles are advected using updated velocities:
\[
\mathbf{x}_p^{n+1} = \mathbf{x}_p^n + \Delta t \mathbf{v}_p^{n+1}.
\]
\section{Additional Details of Experiments}
\label{sec::add_exp}
Our experiments are designed to compare with the most competitive baselines using multiple evaluation metrics, including human evaluation and GPT-based evaluation. Due to page limitations in the main paper, we provide detailed information about the experimental settings, evaluation metrics, and additional results here.

\subsection{Experiments Settings}
\label{sec::design}
\begin{figure}[b]
    \centering
    \includegraphics[width=\columnwidth]{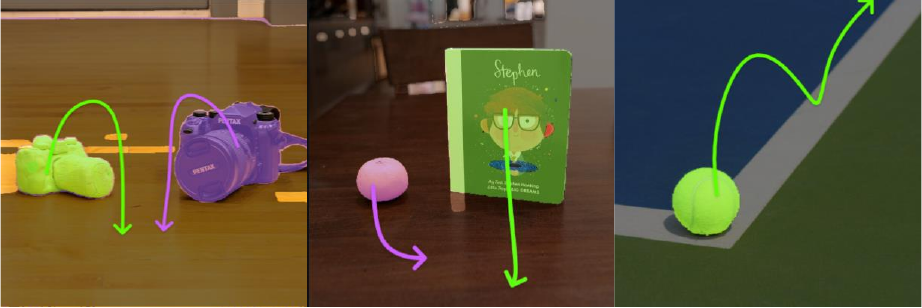}
    \caption{\small
    \textbf{Motion brush input for Kling.} In all cases, we manually define the motion for each object by identifying the movable part and drawing its trajectory. Additionally, we specify the stable part of the object.
    }
    \label{fig:motion_brush}
    \vspace{-5pt}
\end{figure}

In the comparative experiment between our method and baseline generative models, we tried our best to ensure they shared the same generation goal. For our method, we manually assigned an initial 3D velocity to each object. To "interpret" this into text, we described the corresponding dynamics and converted them into prompts such as, \textit{"The elephant hops up and falls onto the ground"} or \textit{"The book falls and the orange rolls forward."} All three baseline models were prompted with the same text. Additionally, Kling supports "motion brush" inputs, which were provided alongside the textual prompt. \Cref{fig:motion_brush} illustrates examples of "motion brush" inputs, where we manually set the stable parts, movable parts, and their trajectory.

\begin{figure*}[t]
    \centering
    \includegraphics[width=\linewidth]{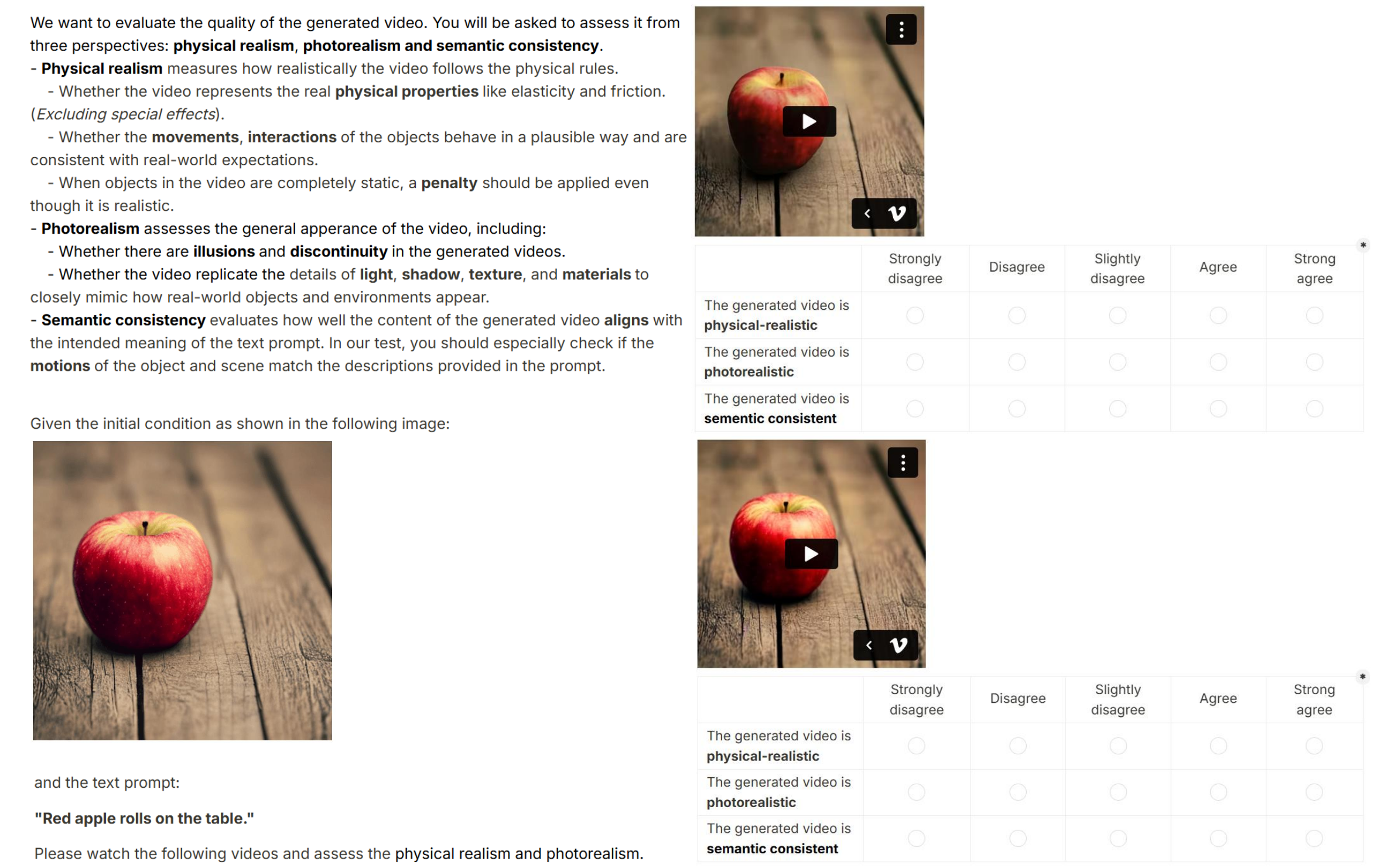}
    \caption{\small
    \textbf{An example page of human evaluation questionnaire.} In each page of the questionnaire, we explain the criteria in detail. We provide the input image, the text prompt and four generated videos in a random order. Each video is followed by a evaluation matrix on a five-point scale, from strongly disagree (1) to strongly agree (5).
    }
    \label{fig:user_study}
    \vspace{-5pt}
\end{figure*}
\begin{figure*}[t]
    \centering
    \includegraphics[width=\linewidth]{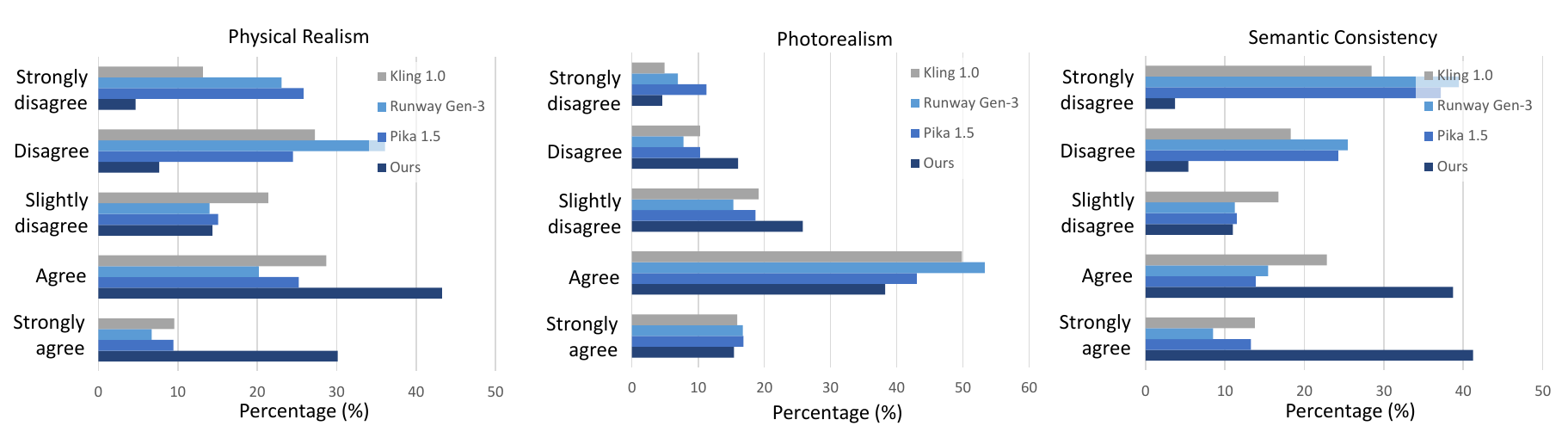}
    \caption{\small
    \textbf{Human evaluation score distribution.} Score distribution shows our method's superiority in physical realism and semantic consistency, with comparable performance across models in photorealism.
    }
    \label{fig:human}
    \vspace{-5pt}
\end{figure*}

\subsection{Evaluation}
In our main paper, we only present the quantitative results of human evaluation. Here, we conduct further experiments using GPT-4 and provide the details.

\textbf{\textit{Human Evaluation.}} We designed a questionnaire to conduct human evaluation, as illustrated in \Cref{fig:user_study}. A total of 31 participants were recruited to complete the 27-page questionnaire. At the beginning, we provided an explanation of video generation models to ensure that participants had a clear understanding of the task. Each page of the questionnaire contains an initial reference image, accompanied by a text prompt describing the expected behavior in the video (e.g., "Red apple rolls on the table"). Four videos are presented on each page in a random order, all corresponding to the same initial condition and text prompt. Participants are instructed to assess each video based on three dimensions. This design ensures a fair, consistent, and comprehensive evaluation process.

\textbf{\textit{GPT-4o Evaluation.}} To assess the quality of the generated videos, we also conducted evaluations using GPT-4o for both our results and the baselines. The prompt is as follows:
\begin{lstlisting}[caption={Prompt used for GPT-4o evaluation}]
I would like you to evaluate the quality of four generated videos based on the following criteria: physical realism, photorealism, and semantic consistency. The evaluation will be based on 10 evenly sampled frames from each video. Given the original image and the following instructions: '{instructions}', please evaluate the quality of each video on the three criteria mentioned above.
Note that: Physical Realism measures how realistically the video follows the physical rules and whether the video represents real physical properties like elasticity and friction. To discourage completely stable video generation, we instruct respondents to penalize such cases. Photorealism assesses the overall visual quality of the video, including the presence of visual artifacts, discontinuities, and how accurately the video replicates details of light, shadow, texture, and materials. Semantic Consistency evaluates how well the content of the generated video aligns with the intended meaning of the text prompt.
Please provide the following details for each video, scores should be ranging from 0-1, with 1 to be the best:
Video 1: Physical Realism Score: [a score]; Photorealism Score: [a score]; Semantic Consistency Score: [a score]
Video 2: Physical Realism Score: [a score]; Photorealism Score: [a score]; Semantic Consistency Score: [a score]
Video 3: Physical Realism Score: [a score]; Photorealism Score: [a score]; Semantic Consistency Score: [a score]
Video 4: Physical Realism Score: [a score]; Photorealism Score: [a score]; Semantic Consistency Score: [a score]
Note that your output should strictly follows the above format, with a ';' after each score. Do not give any other explanations.
The first image is the input image.
# input image
Here are 10 evenly spaced frames from the generated video number {idx + 1}.
# generated frames
\end{lstlisting}

\begin{figure*}[htb]
    \centering
    \includegraphics[width=0.9\linewidth]{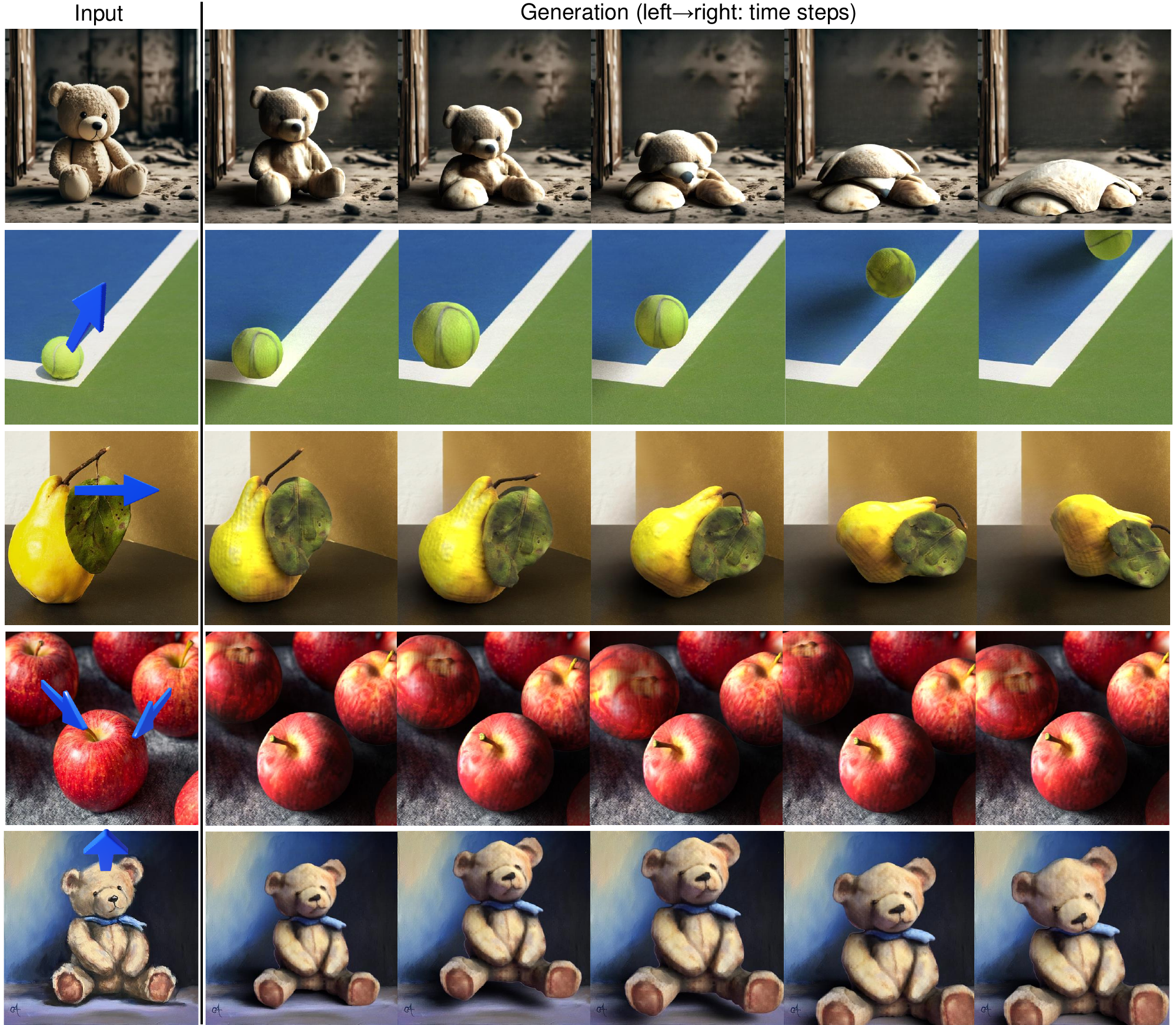}
    \caption{\small
    \textbf{More qualitative results.} The first row demonstrates the \textbf{"sandy" effect}, transforming the teddy bear's material into sand, while the second and third rows showcase \textbf{bouncing} and \textbf{rolling} effects, respectively. The fourth row illustrates a \textbf{multi-object collision} scenario, with three apples colliding with one another, and the final row highlights the system's ability to \textbf{generate a video from a painting}.
    }
    \label{fig:qual_results2}
    \vspace{-5pt}
\end{figure*}
\begin{figure}[!h]
    \centering
        \vspace{-2mm}
        \includegraphics[width=0.45\textwidth]{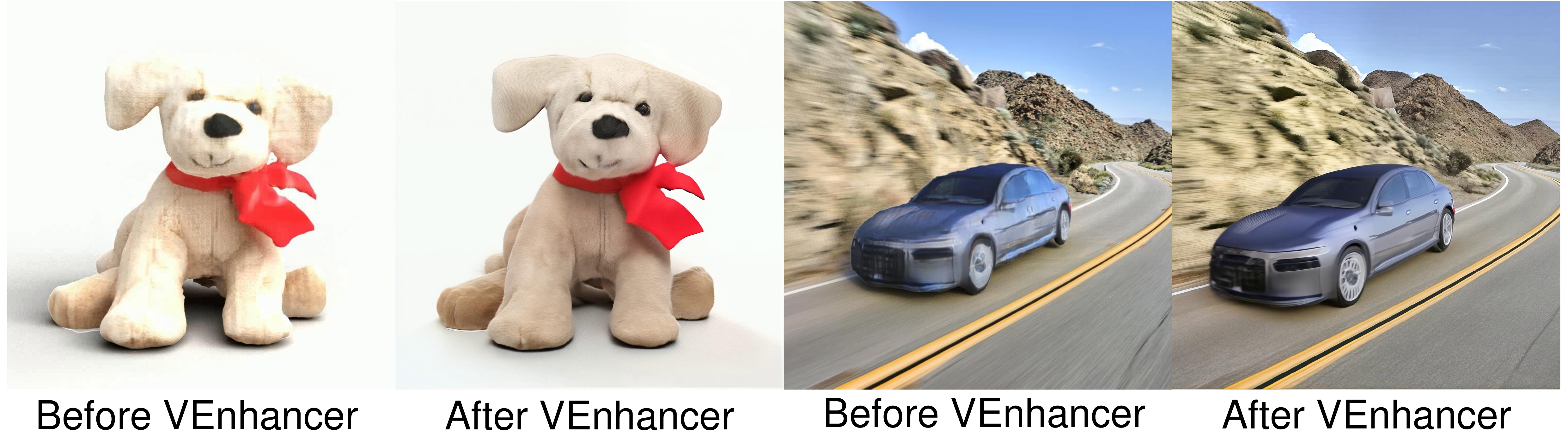}
    \caption{\textbf{Qualitative comparison of VEnhancer.} After post-processing by VEnhancer, more details are recovered and the video appears to be more photorealistic.} 
    \label{fig:venhancer}
\end{figure}
\begin{figure}[!h]
    \centering
        \vspace{-2mm}
        \includegraphics[width=0.45\textwidth]{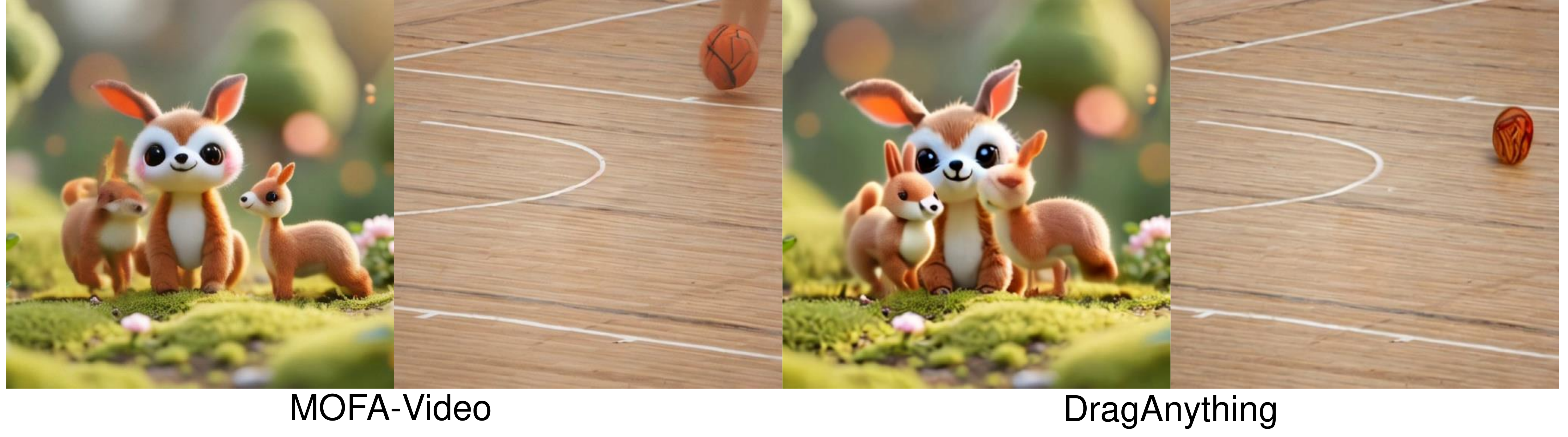}
    \caption{\textbf{Qualitative results of MOFA-Video and DragAnything.} These two open-sourced diffusion models fail to keep background consistent and produce unrealistic deformations.}
    \label{fig:mofa}
\end{figure}

\begin{figure*}[htbp]
    \centering
    \includegraphics[width=0.8\linewidth]{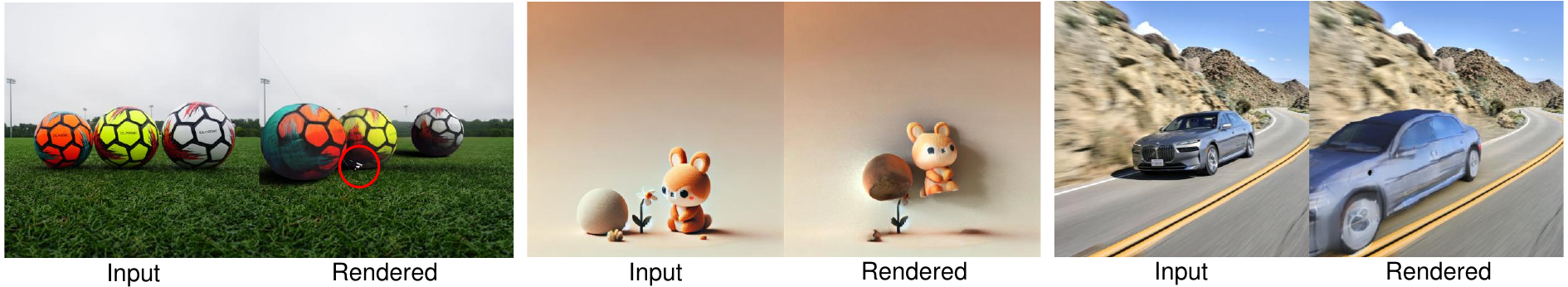}
    \caption{\small
    \textbf{More On Limitations.} The left two images show simulation failures, where unwanted floating points appear in the final rendering results. The middle two images show reconstruction failures, where the wall is recognized as ground by mistake. The right two images depict texture optimization failures, where the car fails to accurately reproduce the real roughness and metallic properties, resulting in an unrealistic appearance.
    }
    \label{fig:limitation_supp}
\end{figure*}
\subsection{Additional Results}
\label{sec::add_results}show that both methods introduce unrealistic deformations. DragAnything sometimes fails to maintain a stable background, even when manually set. MOFA demonstrates better motion control but lacks realism as well. See the table below for quantitative results.
We provide additional quantitative and qualitative results of our experiments.

\textbf{\textit{Human Evaluation Results.}} We analyze the human evaluation scores further in \Cref{fig:human}. The distribution of scores indicates that participants generally agree that most of our results are both physically realistic and semantically consistent. Our method significantly outperforms baseline generative models on these two criteria. However, the four models perform comparably in terms of photorealism.

\textbf{\textit{Additional Qualitative Results.}} Here, we present additional qualitative results in \Cref{fig:qual_results2}. The first row demonstrates the "sandy" effect, where the material of the teddy bear is transformed into sand. The last row illustrates a multi-object collision scenario, where three apples collide with one another. More results are available in video format on our supplementary webpage.

\Cref{fig:venhancer} shows the results after VEnhancer’s post‑processing. Although VEnhancer recovers fine details, it can also introduce hallucinations. This illustrates a fundamental trade‑off between photorealism and physical accuracy: integrating diffusion models into the pipeline leverages their strong priors to compensate for reconstruction and rendering errors, but it cannot guarantee adherence to real‑world physics.

\Cref{fig:mofa} shows the results of two open-sourced diffusion models, MOFA-Video and DragAnything. Both methods introduce unrealistic deformations: DragAnything sometimes fails to maintain a stable background, even when manually set. MOFA demonstrates better motion control but lacks realism as well. Quantitative results of VBench scores in the main paper support these findings. 
\subsection{Applications}
\label{sec::app}
\begin{figure}[!h]
    \centering
        \includegraphics[width=0.45\textwidth]{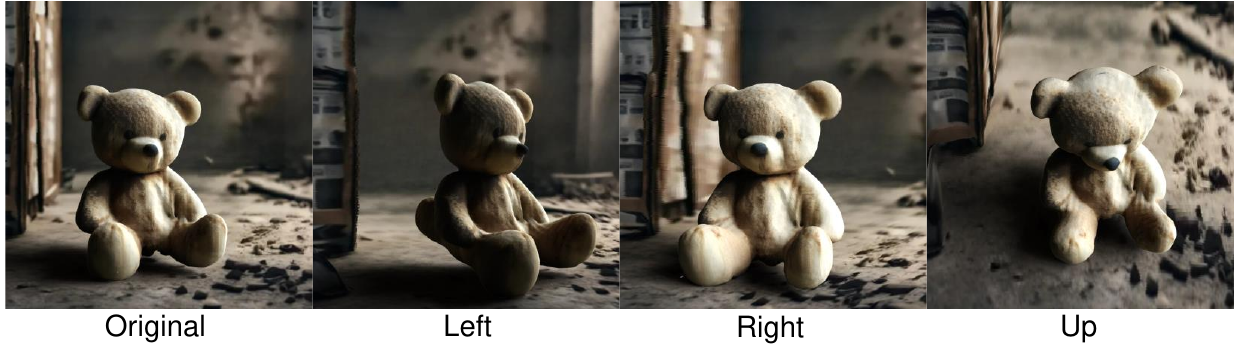}
    \caption{\textbf{Camera controls.} We provide a case demonstrating the potential to perform camera controls on above our pipeline. The left one is the only input image. The right three images are generated with outpaiting and reconstruction.}
    \label{fig:novel}
\end{figure}
Our video generation framework, PhysGen3D, enables a range of exciting applications through its explicit representation. Here are just a few of the compelling use cases our system supports:

\textbf{\textit{Camera controls.}} PhysGen3D's 3D scene representation inherently supports novel view synthesis. We demonstrate this capability (see figure below) by extending our method with minimal modifications: (1) outpainting and meshing the background and (2) rendering from novel views. Results in \Cref{fig:novel} show good consistency across views while maintaining environmental coherence. 

\textbf{\textit{Generate Video from Paintings.}} Thanks to the generalization ability of our interactive 3D world reconstruction pipeline, our method can extend beyond real photos to accommodate other types of inputs, such as generated images and paintings. The final row of \Cref{fig:qual_results2} demonstrates the generation of a video from a painting.

\section{Limitations}
In the main text, we present three failure cases, each highlighting a specific type of error in perception, simulation, and rendering. \Cref{fig:limitation_supp} illustrates additional failures. One involves incorrectly reconstructed meshes with unwanted floating points. Although we have implemented floating point removal during rendering, some points are too close to the object to be detected. Another failure involves material that is incorrectly estimated. The reflectance behavior of cars poses a challenging optimization target, and inaccuracies in inverse rendering result in unrealistic renderings. Failures or inaccuracies may also occur in depth and light estimation. However, these modules are relatively mature, and such errors are comparatively rare.

Many of these failures stem from the inherently ill-posed nature of the task, as reconstructing the full geometry, physics, and textures from partial scene observations requires substantial prior knowledge. 

Currently, we only support a single collider surface, such as the ground or a table. However, our pipeline has the potential to set all stable components as colliders. Additionally, each object is currently homogeneous in density and elasticity. In the future, we may assign different materials to different parts of an object, as demonstrated in \cite{zhai2024physical}.

Overall, our method is designed for object-centric scenes, excelling at mimicking real-world physics for rigid and deformable objects. It also supports a variety of edits and effects. However, reconstructing entire scenes for more complex scenarios remains an open challenge.

\end{document}